\documentclass[10pt]{article}
\usepackage{amsmath}
\usepackage{booktabs}
\usepackage{fancyhdr}

\usepackage[round]{natbib}
\usepackage{geometry}
\usepackage{graphicx}
\usepackage[dvipsnames]{xcolor}
\usepackage{charter}
\usepackage{hyperref}
\hypersetup{colorlinks=true,allcolors=BlueViolet}

\title{\textsc{\textbf{Recurrent Neural Networks for Modelling Gross Primary Production}}}

\author{
David Montero\textsuperscript{1,2,*}\quad 
Miguel D. Mahecha\textsuperscript{1,2,3,4}\quad
Francesco Martinuzzi\textsuperscript{1,4}\\
César Aybar\textsuperscript{5}\quad 
Anne Klosterhalfen\textsuperscript{6}\quad
Alexander Knohl\textsuperscript{6}\\
Franziska Koebsch\textsuperscript{6}\quad
Jesús Anaya\textsuperscript{7}\quad
Sebastian Wieneke\textsuperscript{1} \\
\small\textsuperscript{1}RSC4Earth, IEF, Leipzig University\quad
\textsuperscript{2}iDiv\quad \textsuperscript{3}UFZ\quad
\textsuperscript{4}ScaDS.AI\quad
\textsuperscript{5}IPL, Universitat de Val\`encia\\
\small\textsuperscript{6}Bioclimatology, University of Göttingen\quad
\textsuperscript{7}GEMA, Universidad de Medellín\\
\small*Corresponding author: \texttt{david.montero@uni-leipzig.de}
}
\date{}

\begin{document}

\maketitle

\begin{abstract}
Accurate quantification of Gross Primary Production (GPP) is crucial for understanding terrestrial carbon dynamics. It represents the largest atmosphere-to-land CO$_2$ flux, especially significant for forests.
Eddy Covariance (EC) measurements are widely used for ecosystem-scale GPP quantification but are globally sparse. In areas lacking local EC measurements, remote sensing (RS) data are typically utilised to estimate GPP after statistically relating them to in-situ data.
Deep learning offers novel perspectives, and the potential of recurrent neural network architectures for estimating daily GPP remains underexplored.
This study presents a comparative analysis of three architectures: Recurrent Neural Networks (RNNs), Gated Recurrent Units (GRUs), and Long-Short Term Memory (LSTMs). 
Our findings reveal comparable performance across all models for full-year and growing season predictions. Notably, LSTMs outperform in predicting climate-induced GPP extremes. Furthermore, our analysis highlights the importance of incorporating radiation and RS inputs (optical, temperature, and radar) for accurate GPP predictions, particularly during climate extremes.
\end{abstract}

\thispagestyle{fancy}
\renewcommand{\headrulewidth}{0pt}
\renewcommand{\footrulewidth}{0pt}
\fancyhf{}
\cfoot{\scriptsize{Copyright 2024 IEEE. Published in the 2024 IEEE International Geoscience and Remote Sensing Symposium (IGARSS 2024), scheduled for 7 - 12 July, 2024 in Athens, Greece. Personal use of this material is permitted. However, permission to reprint/republish this material for advertising or promotional purposes or for creating new collective works for resale or redistribution to servers or lists, or to reuse any copyrighted component of this work in other works, must be obtained from the IEEE. Contact: Manager, Copyrights and Permissions / IEEE Service Center / 445 Hoes Lane / P.O. Box 1331 / Piscataway, NJ 08855-1331, USA. Telephone: + Intl. 908-562-3966.}}

\section{Introduction}
\label{sec:intro}

Terrestrial ecosystems are pivotal in carbon sequestration \citep{Friedlingstein2023globalcarbonbudget}, with forests contributing significantly to this process \citep{Bonan2008}. The single largest flux in this context is the gross uptake of CO$_2$ by vegetation ``Gross Primary Production'' (GPP). Hence, accurately quantifying forest GPP is essential for monitoring terrestrial carbon dynamics. Traditionally, GPP at the ecosystem scale is estimated using in-situ measurements of net CO$_2$ exchange (NEE) obtained through Eddy Covariance (EC) systems. Despite the growing deployment of EC measurements within global networks (e.g. Fluxnet, \citealp{Baldocchi2019}) , which are built on regional or continental initiatives (e.g. ICOS, Asiaflux, Ameriflux), these networks still exhibit a sparse global distribution \citep{Schimel2015}. Remote sensing data, particularly Vegetation Indices (VIs), are commonly employed as weak proxies for GPP estimation where no local EC estimations are performed \citep{Zeng2022}.
Additionally, integrating Artificial Intelligence (AI) methods with remote sensing data, sometimes in conjunction with climate data, has facilitated GPP product generation \citep{Jung2019}. However, while recurrent neural network architectures, which exploit temporal dependencies, have been used for modelling VIs \citep{Martinuzzi2023} and land-atmosphere interactions \citep{Besnard2019}, their potential for modelling daily forest GPP remains underexplored.

This study conducts a comparative analysis of three recurrent architectures: 1) Recurrent Neural Networks, RNNs, 2) Gated Recurrent Units, GRUs, and 3) Long-Short Term Memory networks, LSTMs. Additionally, we explore the models' predictive performance under climate-induced GPP extreme values and provide insights into the importance of the utilised features. The paper is organised as follows: Sec.~\ref{sec:methods} details the data preprocessing and AI approaches, Sec.~\ref{sec:results} presents and discusses the results, and Sec.~\ref{sec:conclusions} outlines the conclusions.

\section{Methods}
\label{sec:methods}

\subsection{Data preparation}
\label{subsec:data}

The study period spanned from 2016 to 2020, aligning with the availability of Sentinel-2 (S2) data and the ICOS 2020 Warm Winter dataset \citep{icos2022warmwinter}. The data preprocessing procedures were conducted as follows:

\textbf{Gross Primary Production}. Daily GPP data were sourced from the ICOS 2020 Warm Winter dataset. Forest cover information was derived from CORINE 2018 Land Cover (CLC2018), and sites with less than 70\% forest cover within a 1 km radius of the EC tower were excluded. Sites with urban development within the buffer and locations outside Europe were excluded. Daily GPP values (g C m$^{-2}$ d$^{-1}$) for each site were obtained from the nighttime partitioning method \citep[\texttt{GPP\_NT\_VUT\_REF}, ][]{Reichstein2005nightpartition}. 
Timesteps with less than 70\% high-quality measurements (using \texttt{NEE\_VUT\_REF\_QC}) and GPP values below zero were discarded. Additionally, sites with less than 60\% valid data during the study period were excluded. A total of 19 forest sites, categorised as 12 Evergreen Needleleaf Forests (ENF), 4 Deciduous Broadleaf Forests (DBF), and 3 Mixed Forests (MF), were retained for analysis.  Values below the 10\% threshold of the negative tail of the anomalies distribution were flagged. Anomalies were computed as the deviation from the raw time series to the mean seasonal cycle, calculated from all available years in the dataset. Segments of 5 or more consecutive days with flagged anomalies were identified as climate-induced GPP extremes.

\textbf{Sentinel-2}. S2 L2A data cubes centred on each EC tower location were created. Non-10 m bands were resampled to this resolution via nearest neighbours. Cloud and cloud shadow pixels were masked using a CloudSEN12-based model \citep{Aybar2022cloudsen12}, while snow pixels were masked with Fmask v3.2. Surface reflectances underwent conversion to Nadir BRDF Adjusted Reflectance (NBAR) using the c-factor method \citep{Roy2017a}. The CLC2018 was employed to mask out non-forest pixels. Timesteps with less than 1\% clear surface pixels within the 1 km buffer were excluded, and NBAR values were spatially averaged within it. A total of 122 VIs from Awesome Spectral Indices v0.5.0 \citep{Montero2023asi}, designed for S2, were computed using the NBAR values. To manage the substantial correlation among VIs and ensure interpretability, Principal Component Analysis (PCA) was performed. The first 18 Principal Components (PCs), jointly explaining $>$99\% of the variance, were selected as predictors in the GPP modelling. S2 features are denoted as S2$_i$, where $i$ is one of the PCs.

\textbf{Sentinel-1}. Sentinel-1 (S1) data cubes with a pixel size of 10 m and centred on each EC tower were created. We used the Radiometrically Terrain Corrected (RTC) product, which consists of terrain-flattened $\gamma^0$ backscattering values. Both $\gamma^0$ values ($\gamma^0_{\textrm{VV}}$ and $\gamma^0_{\textrm{VH}}$) were used. The Dual Polarised Radar Vegetation Index for S1 (DpRVI$_{\textrm{VV}}$) was computed from the $\gamma^0$ values in linear units. $\gamma^0$ values were also scaled to dB units. The CLC2018 was employed to mask out non-forest pixels. All features were spatially averaged within 50 and 1,000 m from each EC tower, aimed at minimising potential signal contributions from the tower itself. S1 features are denoted as S1$_i$, where $i$ is one of the $\gamma^0$ values or the DpRVI$_{\textrm{VV}}$.

\textbf{Land Surface Temperature (LST)}. MODIS daily LST v6.1 data were obtained from Terra (MOD11A1) and Aqua (MYD11A1). LST values were spatially smoothed using a mean $3\times3$ kernel. The pixel values intersecting the coordinates of each EC tower were retrieved. LST features are denoted as MOD11A1$_i$ or MYD11A1$_i$, where $i$ is one of the daytime (dt) or nightime (nt) LST values.

\textbf{Simulated clear-sky radiation}. The daily mean clear-sky radiation (W m$^{-2}$) was simulated individually for each site \citep{reda2008solar}.

\subsection{Modelling and evaluation approach}
\label{subsec:modelling}

We interpolated missing data in the predictor features linearly, while we did not interpolate GPP values, even when missing. A temporal split was implemented, utilising 2016-2018 as the training set and 2019-2020 as the test set. The three models (RNNs, GRUs, and LSTMs) were hyper-tuned using HyperBand, employing a sequence-to-one approach with a sequence length of 90 days. The hyperparameter configuration included a maximum of 5 layers and a maximum of 512 units per layer. Dropout layers with a 20\% dropout rate were incorporated. Weights were initialised by sampling from a random uniform distribution within $[-\sqrt{n},\sqrt{n}]$, where $n$ is the number of features. The Adam optimiser was used, and the learning rate was hyper-tuned from 0.0001 to 0.01 on a logarithmic scale. The Mean Absolute Error (MAE) served as the loss function. Following hyperparameter tuning, the best configurations were employed, and the models were trained for 300 epochs, saving the checkpoint for the best performance on the test set.

All models were evaluated using the Normalised Root Mean Squared Error (NRMSE), calculated as $\textrm{NRMSE}=\textrm{RMSE}/(y_{\textrm{max}}-y_{\textrm{min}})$, where RMSE is the Root Mean Squared Error, and $y_{\textrm{max}}$ and $y_{\textrm{min}}$ indicate the maximum and minimum observed values. This evaluation was performed on the test set using three different settings: 1) full period, 2) growing season (May-Sep), and 3) climate-induced GPP extremes. 

\begin{figure}[!t]
    \centering
    \includegraphics[width=0.8\textwidth]{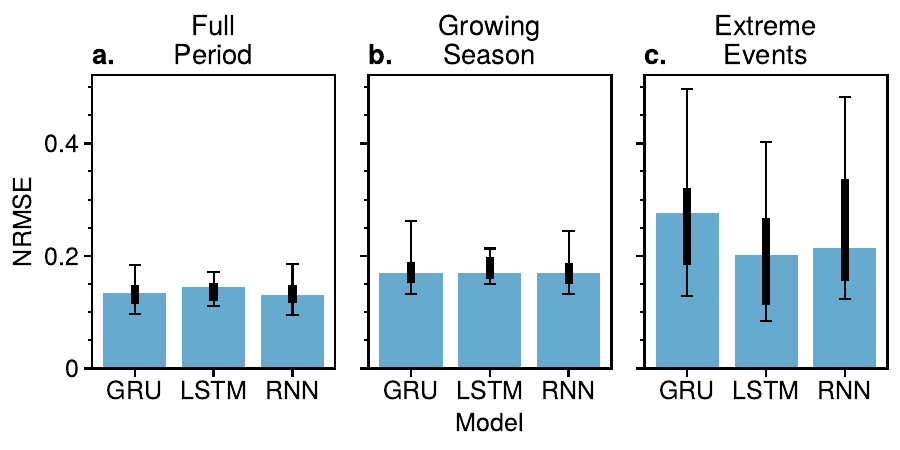}
    \caption{\textbf{Comparison of models performance}. Median NRMSE per model for a) full period, b) growing season, and c) climate-induced GPP extremes. Black boxes represent the range from Q$_1$ to Q$_3$ while error bars denote the range between the 5th and 95th percentiles.}
    \label{fig:model-comparison}
\end{figure}

\subsection{XAI: Feature Importance}
\label{subsec:xai}

Feature importance (FI) for a specific feature $f$ was calculated as $\textrm{FI}_f=E_f-E_b$, where $E_f$ is the NRMSE of predictions derived from a permuted version of the test set, where the feature array was randomly shuffled; and $E_b$ is the baseline NRMSE of predictions derived from the non-permuted test set. This process was iterated ten times, and the mean was calculated per site.

\section{Results and Discussions}
\label{sec:results}

\subsection{Comparison of models}
\label{subsec:comparison}

The three models exhibit comparable predictive performance throughout the entire test period (Fig.~\ref{fig:model-comparison}a). The NRMSE increases for all models when considering predictions solely during the growing season, with comparable outcomes. LSTMs consistently maintain the lowest error variance (Fig.~\ref{fig:model-comparison}b). Upon examining predictions related to climate-induced GPP extremes, all models exhibit increased NRMSE and a higher error variance. However, this increase is comparatively lower for LSTMs, followed by RNNs, while GRUs show the most significant escalation among the three models (Fig.~\ref{fig:model-comparison}c).

\begin{figure*}[!t]
    \centering
    \includegraphics[width=1\textwidth]{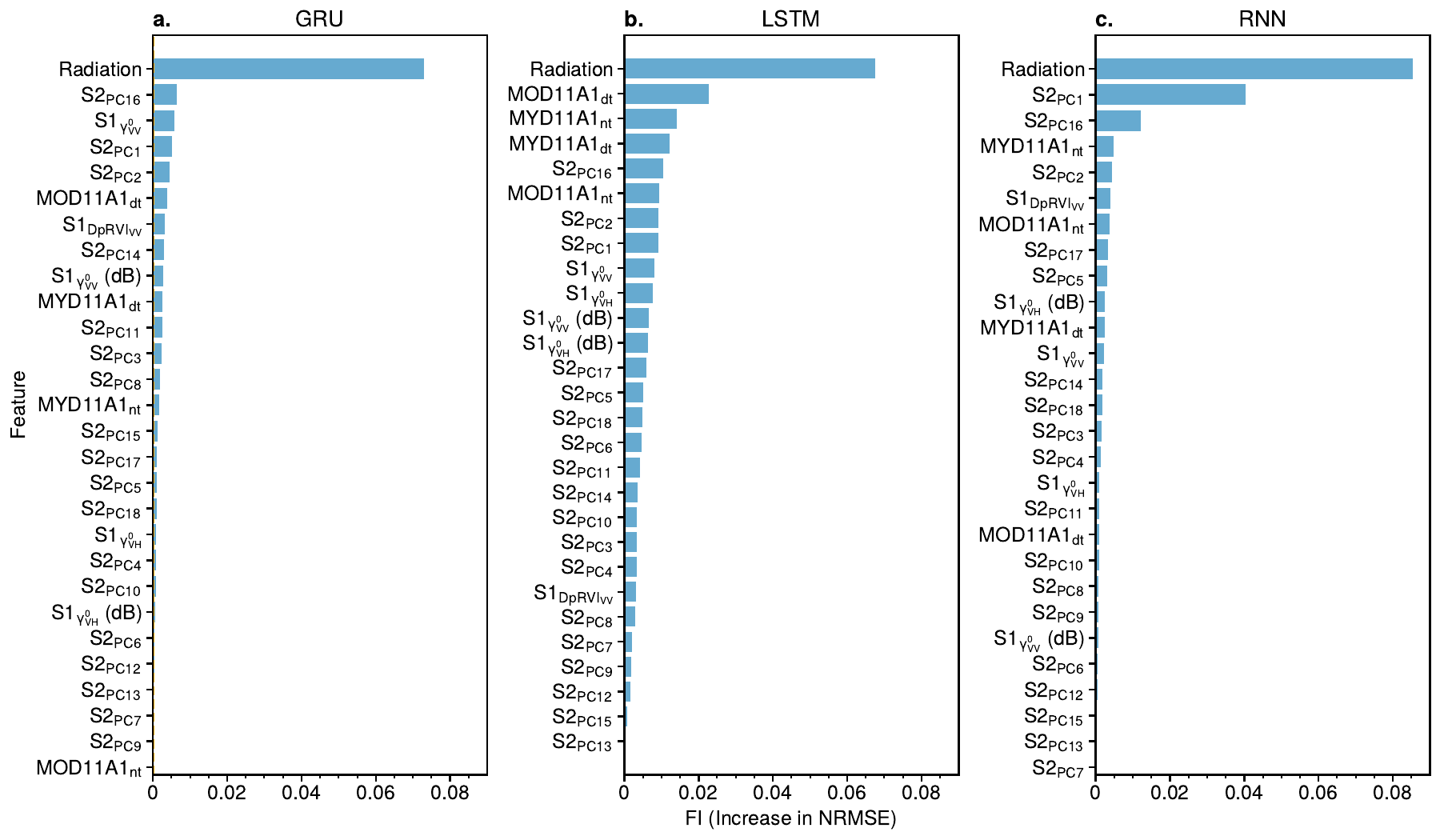}
    \caption{\textbf{Feature Importances (FIs) per model}. Median FIs (expressed in increased units of NRMSE) for a) GRUs, b) LSTMs, and c) RNNs. Features are sorted in decreasing order by model.}
    \label{fig:fi}
\end{figure*}

\subsection{Feature Importance}
\label{subsec:fi}

\begin{figure}[!t]

    \centering
    \includegraphics[width=0.9\textwidth]{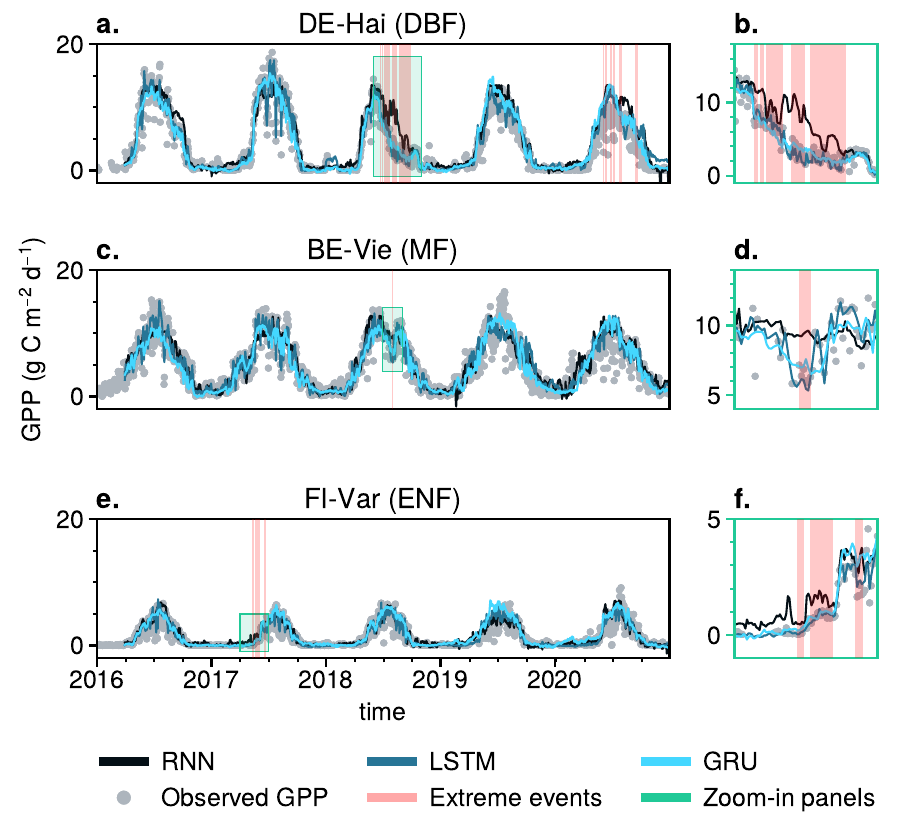}
    \caption{\textbf{GPP predictions for three example sites}. The first column shows predictions for a) DE-Hai (DBF), c) BE-Vie (MF), and e) FI-Var (ENF). The second column displays zoom-in panels depicting extreme events at each site.}
    \label{fig:example-sites}

\end{figure}

Notably, the simulated clear-sky radiation is a pivotal factor for all models, evident from its highest FI ranking (Fig.~\ref{fig:fi}). This is attributed to its role in providing site-specific physical radiation values crucial for modulating the GPP prediction range, as illustrated by predictions across example sites characterised by variations in latitude and forest type (Fig.~\ref{fig:example-sites}). Following radiation, S2$_{\textrm{PC16}}$, S2$_{\textrm{PC1}}$, and S2$_{\textrm{PC2}}$ exhibit notably higher FIs in GRUs and RNNs compared to other features. S2$_{\textrm{PC16}}$ is primarily linked to chlorophyll-related VIs (e.g. Modified Chlorophyll Absorption in Reflectance Index, MCARI) with a slight association with some water-related VIs (e.g. Disease-Water Stress Index, DSWI). Meanwhile, S2$_{\textrm{PC1}}$ is highly correlated with productivity-related VIs (e.g. Kernel Normalised Difference Vegetation Index, kNDVI), and S2$_{\textrm{PC2}}$ is closely associated with water-related VIs (e.g. Normalised Difference Moisture Index, NDMI). In the case of LSTMs, these S2 features gain importance shortly after most LST-based features. We speculate that this FI explains the higher predictive performance of LSTMs during extreme events, where LST plays a more pivotal role than optical-based features for forests, especially during climate extremes like droughts \citep{HoekvanDijke2023forest_grasslands_lst_optical}. Following these features, in LSTMs, the significance of most S1 features is notable, with FIs surpassing those of the remaining S2 features (a distinction less clear in RNNs and GRUs).

\section{Conclusions}
\label{sec:conclusions}

We assessed recurrent neural networks for daily GPP modelling using remote sensing data. Our key observation is that all models perform similarly in predicting daily GPP over the growing season or the entire year. However, errors escalate when predicting climate-induced GPP extremes. Here, LSTMs demonstrate a slightly lower error rate in predicting climate-induced GPP extremes. Moreover, our analysis underscores the importance of simulated clear-sky radiation in modulating the GPP response. Additionally, we find that LST data proves to be effective for predicting climate-induced GPP extremes, complementing the information derived from optical and radar data obtained from S2 and S1. 

\bibliographystyle{abbrvnat}
\bibliography{Bib}

\end{document}